\documentclass{article} 
\usepackage{iclr2016_conference,times}
\usepackage{hyperref}
\usepackage{url}
\usepackage{graphicx} 
\usepackage{tabularx}
\usepackage{makecell}
\usepackage{amsmath}
\usepackage{amssymb}
\usepackage{footnote}
\usepackage{colortbl}
\usepackage{diagbox}
\usepackage{amsmath}
\usepackage{blindtext}

\newcommand{\BigO}[1]{\ensuremath{\operatorname{O}\bigl(#1\bigr)}}

\definecolor{LimeGreen}{rgb}{0.1,0.9,0.1}
\definecolor{Maroon}{rgb}{0.9,0.1,0.1}
\newcommand{\no}{ $\color{Maroon}\times$ }
\newcommand{\yes}{ $\color{LimeGreen}\checkmark$  }

\def\works{\cellcolor{green!50}Works}
\def\fails{\cellcolor{red!50}Fails}
\def\struggles{\cellcolor{yellow!50}Struggles}

\newcommand{\fig}[1]{Fig.~\ref{fig:#1}}
\newcommand{\tab}[1]{Table~\ref{tab:#1}}

\title{Extensions and Limitations of the Neural GPU}

\author{Eric Price\thanks{Work done while at OpenAI.} \\
The University of Texas at Austin\\
\texttt{ecprice@cs.utexas.edu} \\
\And
Wojciech Zaremba, Ilya Sutskever\\
OpenAI\\
\texttt{\{woj,ilyasu\}@openai.com} \\
}

\makeatletter
\let\@fnsymbol\@arabic
\makeatother

\begin{document}

\maketitle

\begin{abstract}

The Neural GPU is a recent model that can learn algorithms such as multi-digit binary addition and binary multiplication in a way that generalizes to inputs of arbitrary length.  We show that there are two simple ways of improving the performance of the Neural GPU: by carefully designing a curriculum, and by increasing model size.  The latter requires a memory efficient implementation, as a naive implementation of the Neural GPU is memory intensive.  We find that these techniques increase the set of algorithmic problems that can be solved by the Neural GPU: we have been able to learn to perform all the arithmetic operations (and generalize to arbitrarily long numbers) when the arguments are given in the decimal representation (which, surprisingly, has not been possible before). We have also been able to train the Neural GPU to evaluate long arithmetic expressions with multiple operands that require respecting the  precedence order of the operands, although these have succeeded only in their binary representation, and not with perfect accuracy.

In addition, we gain insight into the Neural GPU by investigating its failure modes.  We find that Neural GPUs that correctly generalize to arbitrarily long numbers still fail to compute the correct answer on highly-symmetric, atypical inputs: for  example, a Neural GPU that achieves near-perfect generalization on decimal multiplication of up to 100-digit long numbers can fail on $000000\dots002 \times 000000\dots002$ while succeeding at $2 \times 2$.  These failure modes are reminiscent of adversarial examples.

\end{abstract}

\section{Introduction}

Deep Neural Networks (DNNs) are extremely useful for solving difficult
pattern recognition tasks for two reasons: first, because they can
compactly represent good solutions to difficult pattern recognition
tasks; and second, because these good solutions can be found with
stochastic gradient descent.  It is not immediately obvious that a DNN
should have the ability to represent solutions to such problems compactly.
This is the case because the depth and width of DNNs allows them to
simulate any parallel computer that runs for a modest number of
steps, making it possible for the DNN to match the performance of
\emph{any} parallelizeable statistical ML model by simply simulating
it.  This is one of the reasons DNNs are successful relative to
other ML models.

DNNs are especially useful in the supervised learning setting, where
the goal is achieve low test error over a given data distribution.
Statistical learning theory \citep{book} guarantees that this can be
done by minimizing training error, as long as the training data is
drawn from the same data distribution and when there is more training
cases than parameters.

While the ability to achieve low test error on a specific data
distribution is extremely useful in practice and has already enabled a
wide range of practical applications, there is evidence that this
ability does not fully capture the intuitive notion of
pattern recognition.  For example, the existence of adversarial
examples, which are data cases that are nearly indistinguishable from
real data cases that confuse all existing discriminative classifiers,
suggests that supervised DNNs are substantially less robust than human
pattern recognition. And indeed, we would expect a system that has fully
``understood'' the relevant visual (say) concepts to not be fooled by
adversarial examples.  Understanding and fixing this problem is an
active area of research.

Another domain where a mere low error on a specific data distribution
seems unsatisfying is the domain of simple algorithms.  Simple
algorithms have a well-defined output for all conceivable inputs, so
if we collect enough of input-output examples from a reasonably-chosen
distribution where the outputs are computed by some (unknown) simple
algorithm, a sufficiently good learning method ought to be able to
infer the ``true'' algorithm: one that can perfectly generalize to all
conceivable inputs, and not just the inputs that tend to occur in the
training data distribution.

This problem lies at the core of \emph{program induction}, an old
field that has significant past work \citep{nordin1997evolutionary,
  liang2013learning, wineberg1994representation, solomonoff1964formal,
  Holland,Goldberg, gomez2008accelerated}.  More recently, researchers
have begun investigating this problem using the deep learning techniques of
neural network function approximation and stochastic gradient descent
\citep{graves2014neural,zaremba2015reinforcement,KS15,
  kurach2015neural, andrychowicz2016learning}. All these works have
been able to learn algorithms that correctly generalize to inputs of
much greater length than the training cases on some problems.  The
Neural GPU \citep{KS15} is notable among these since it is the only
model that has, thus far, been able to learn to correctly multiply
integers of length much greater than it has been trained on.

The phenomenon of generalization to inputs that are outside the
training distribution is poorly understood.  The space of problems for
which such generalization is possible has not been identified, and a
detailed understanding of the causes of such generalization are
missing as well.  Given that the test inputs do not come from the
training data distribution, we do not yet have a formal reason to believe
that such out-of-distribution generalization should succeed. 

In this paper, we attempt to better understand this generalization in
the context of the Neural GPU.  We empirically study the parameters
that affect its probability of successfully generalizing to inputs
much greater length, and also study its failures.  We report three
notable findings: first, that larger models can generalize on harder
tasks; second, that very detailed curriculum can enable the training
of otherwise un-trainable neural networks; and third, those models
that achieve perfect generalization on longer test cases when they are
drawn from the uniform distribution still fail on highly structured
inputs.  This suggests that these models fail to learn the ``true
algorithm'' as well as we've hoped, and that additional research is
required for to learn models that can generalize much better.

The code for this work can be found at
\url{https://github.com/openai/ecprice-neural-gpu}.

\section{Related Work}
 
The problem of learning algorithms from data has been investigated in
the field of program synthesis~\citep{nordin1997evolutionary,
  liang2013learning, wineberg1994representation,solomonoff1964formal}
and genetic programming~\citep{Holland,Goldberg,gomez2008accelerated}.
These approaches usually aim to directly produce the source code of an
algorithm that solves the problem specified by the training data.

A recent approach to algorithm learning attempts to use the power of
neural networks and their learning algorithm.  Neural networks are
inherently robust, and are capable of dealing with ``imprecise'' data
(such as images or text) that can be difficult to handle in models that
directly work with source code.  There exist a number of neural network
architecture that implement this idea: the Neural Turing Machine
(NTM)~\citep{graves2014neural}, 
the standard LSTM (to some extent)~\citep{zaremba2014learning}, the Grid
LSTM~\citep{kalchbrenner2015grid}, the Stack
RNN~\citep{joulin2015inferring}, the Neural
DeQue~\citep{grefenstette2015learning}, the End-to-End Memory
Networks~\citep{sukhbaatar2015weakly}, the Hierarchical Attentive
Memory~\citep{andrychowicz2016learning}, and the Neural random-access
machines~\citep{kurach2015neural}.

For a neural model to be able to learn an algorithm, it is essential
that it is capable of running the necessary number of computational
steps.  Most of the above models have only been successfully used to
learn algorithms that require a linear number of computational steps
in the size of the input.  While some
models~\citep{zaremba2015reinforcement, zaremba2015learning,
  graves2016adaptive} can in principle learn the correct runtime
for a given algorithm, in practice it has not been possible to learn
algorithms requiring superlinear runtime, such as long integer
multiplication.  The only known neural model that can solve tasks
whose runtime is truly superlinear in the size of the input is the
Neural GPU~\citep{KS15}, which is the model that we investigate
further in this paper.

The Grid LSTM~\citep{kalchbrenner2015grid} is a powerful architecture
that has been used to successfully learn 15-digit decimal addition.
This model is similar to the Neural GPU -- the main difference is that
the Neural GPU is less recurrent.  The Neural GPU has been shown to be
able to generalize outside its training distribution, and while this
has not been shown for the Grid LSTM, we believe that with appropriate
modification, it should be capable of such generalization as well.

\begin{figure}[h]
  \centering
  \begin{minipage}{.48\linewidth}
    \includegraphics[width=\linewidth]{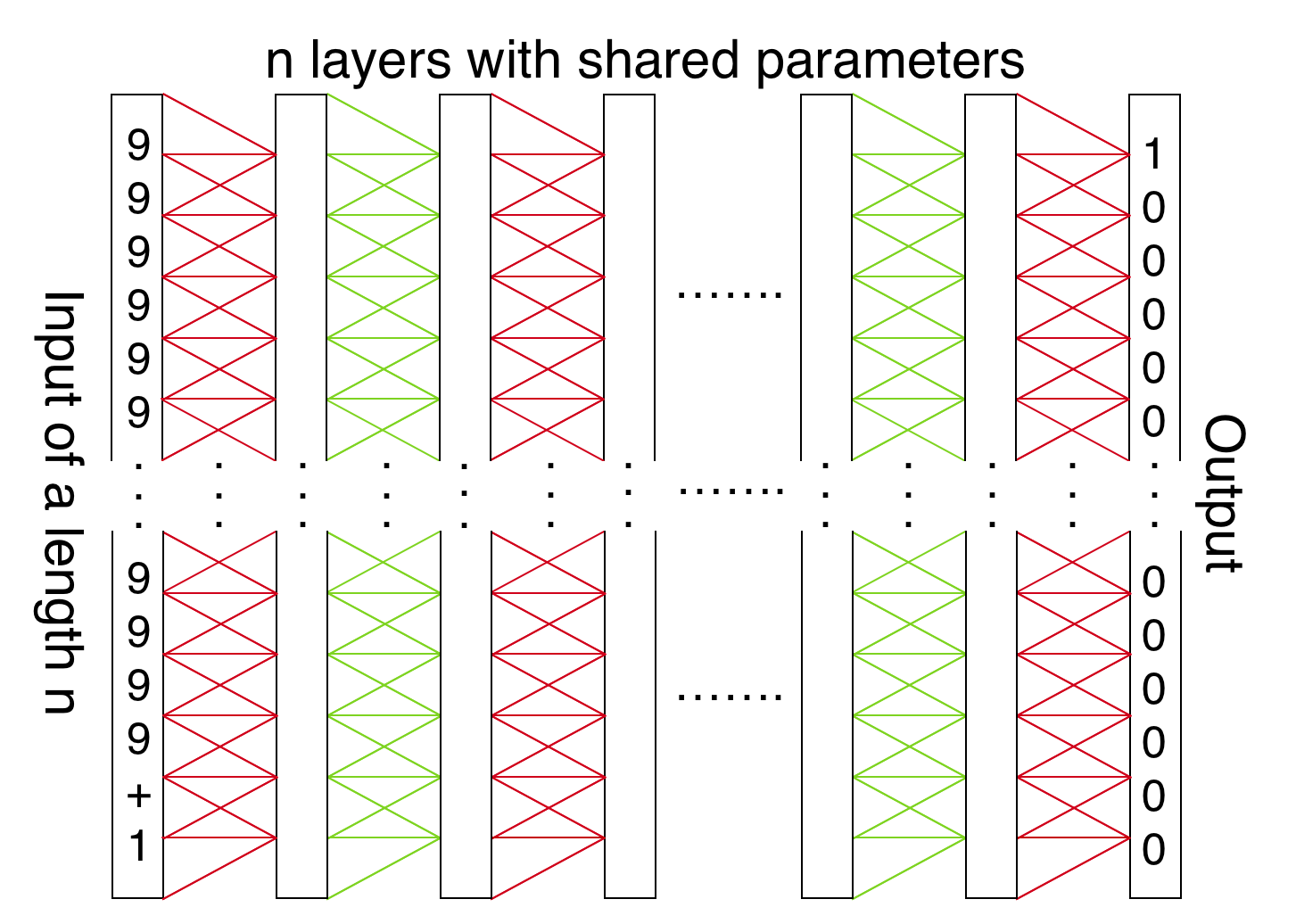} 
  \end{minipage}
  \begin{minipage}{.48\linewidth}
    The red color denotes shared convolutional filters, and green
    denotes a different set of shared convolution filters. Convolution
    is applied to the variable length input, and the weights of every
    other layer are shared. This architecture has fixed number of
    parameters for variable length input, and performs quadratic
    computation in its length.
  \end{minipage}
  \caption{The Neural GPU.}
  \label{fig:model}
\end{figure}

Neural network models that are capable of learning and representing
algorithms tend to be extremely deep and have an elaborate
architecture, which makes the problem of minimizing their training
error to be very challenging for stochastic gradient descent.  To
reduce the difficulty of the training problem, curriculum learning has
been found to be critical. In particular, all the aforementioned
models require a curriculum learning~\citep{bengio2009curriculum,
  zaremba2014learning} in order to successfully learn sophisticated
functions, and  the results in this paper are no different.

\section{Model}

The Neural GPU is an architecture developed by \cite{KS15} that
can learn algorithms such as binary multi-digit addition and
multi-digit multiplication. The main idea of the Neural GPU is to
build an architecture capable of performing the necessary computation
for the task without being overly deep. By being less deep, the
problem of minimizing its training error becomes easier.  Tasks that
require a super-linear number of computational operations (in the size of the
input) cannot be solved by a neural architecture that can only perform
a linear number computational operation.  \tab{complexity} lists the number of
computational operations that can be performed by several different neural
network architectures. In our notation, a feed forward network
consumes input of a predefined, constant size $\BigO{1}$, and performs
a fixed number of computation operations (also $\BigO{1}$).  Other architectures,
such as classical convolution networks, also have a predefined input size
($\BigO{1}$), and process it with a constant $\BigO{1}$ number of computational operations.
However, it is possible to apply the convolution operation to an input
of variable size. This approach is sometimes used in object detection,
where the same convolutional neural network is applied on images of
variable size~\citep{sermanet2013overfeat}. Similarly, the recurrent
neural network (RNN) is capable of processing inputs of variable
length, where the amount of computation performed by the RNN is linear
in the length of the sequence.

\begin{table}[h]
\tiny
\centering
\renewcommand{\arraystretch}{1.1}
\begin{tabular}{|c|c|c|c|c|c|c|c|}
\hline 
                         & Feed forward & Classical & CNN & RNN & Neural Turing & Grid LSTM & Neural GPU \\ 
                         & network    & CNN & for detection& & Machine & & \\ \hline
Input Size               &    $\BigO{1}$        &     $\BigO{1}$                &   $\BigO{n}$                    &  $\BigO{n}$  & $\BigO{n}$    & $\BigO{n}$     & $\BigO{n}$ \\ 
Number of steps   &    $\BigO{1}$        &     $\BigO{1}$                &   $\BigO{n}$                    &  $\BigO{n}$  & $\BigO{n}$    & $\BigO{n^2}$    & $\BigO{n^2}$ \\ \hline
\end{tabular}
\caption{This table compares the number of computational steps
  performed by various architectures as a function of their input
  size. In our notation, $\BigO{1}$ means that the number of
  steps (or the size of the input) are constant.  In contrast,
  $\BigO{f(n)}$ means that the input can be of variable length $n$, and that the
  number of steps grows as $f(n)$ in the size of the input.}
\label{tab:complexity}
\end{table}

The Neural GPU architecture is the combination of a convolution on
variable size inputs with a recurrent neural network as shown in \fig{model}.  The
Neural GPU consumes an input of a variable length $n$. It repeatedly
applies several convolutional layers $n$ times, where $n$ is the
length of the input; thus, the depth of the model is dependent on the
length of the input.  It performs $\BigO{n^2}$ operations for each
input of the length $\BigO{n}$. This is a desirable property, because
now there is a possibility that the model could learn to represent
algorithms whose runtime grows superlinearly in the size of the input, such as integer
multiplication.  Harder problems can require significantly more computational
steps.  While we could, in principle, unroll our models for an
enormous number of timesteps to make them capable of solving even
NP-hard problems, it is exceedingly unlikely that gradient learning
would succeed training models of such depth.

The architecture most similar to Neural GPU is the Grid LSTM
\citep{kalchbrenner2015grid}. It has been shown to learn 15 digit
long decimal addition task, although it has not yet been shown to
generalize to inputs of length greater than the training data.

To successfully train the Neural GPU, \cite{KS15} used the following
techniques:
\begin{itemize}
    \item The architecture is that of a gated recurrent unit (GRU) through depth~\citep{bahdanau2014neural}.
    \item Tanh cutoff: the hyperbolic tangent activations used by the GRU
      are truncated once they reach a critical upper (and lower) bound.
      The hope is that this makes the results more ``digital''.
    \item The use of Dropout~\citep{dropout}.
    \item Instead of using the same weights at every pair of layers,
      the Neural GPU starts out by cycling through 6 independent sets
      of weights, which are gradually annealed to become identical as
      optimization progresses.
    \item Gradient noise: the original paper notes it as important,
      but the released code has this option disabled.
\end{itemize}
In this paper, we use all these techniques and analyze their
importance \fig{hyper}.  More details about the meaning of each
modification are available in the original work~\citep{KS15}.

\section{Improvements to the Neural GPU}

The performance obtained by the Neural GPU is very sensitive to the
random seed used to produce the initialization and the order of the
training data.  For example, when we ran several identical models with
different random seeds, we find, similarly to \cite{KS15}, that only a
fraction of them generalizes to over $99\%$ cases (where the test
cases are of length much greater than the training cases, and the
model is considered to have made an error when it mispredicts even one
bit of the output vector).  Such variability makes it hard to
hill-climb in hyperparameter space.  We address this problem by
running $\approx 15$ instances of the each model with different seeds
on each of our $13$ tasks.  We measure the fraction of seeds that
cause the model to surpass $99\%$ accuracy on the (longer and out of
distribution) test data. This fraction provides us with a reliable
estimate of the quality of the hyperparameter configuration. Moreover, 
such measurements had to be done on several problems. Particular 
``tricks'' often are effective just for a single problem, and damage performance on 
others.

\fig{hyper} shows the success rate for a given size and model
modification.  We find that many model modifications help reduce
training error but do not with generalization. Note that most changes
to the model increase the success rate by no more than $10\%$, which is
too small to be detected without the thorough experimental framework
that we've outlined above.

\begin{figure}[!h]
    \tiny
    \centering
    \renewcommand{\arraystretch}{1.1}
    \begin{tabular}{|l||c|c|c||c|c|c|}
      \hline
                                   & \multicolumn{3}{c||}{{\bf Training data}} & \multicolumn{3}{c|}{{\bf Test data}} \\
      \hline
      \diagbox{Model}{\# Filters}  & 24  & 128  & 256 & 24 & 128 & 256 \\
      \hline
      \hline
      Classical Neural GPU                        & $70\%$ & $99\%$   & $100\%$ & $41\%$ & $53\%$ & $53\%$ \\
      without tanh cutoff                         & $58\%$ & $99\%$   & & $29\%$ & $45\%$  & \\
      without dropout                             & $83\%$ & $100\%$  & & $35\%$ & $46\%$  & \\
      one independent copy of weights             & $67\%$ & $77\%$   & & $44\%$ & $62\%$  & \\
      with noise on gradients                     & $72\%$ & $100\%$  & & $43\%$ & $53\%$  & \\
      with batch-normalization                    & $85\%$ & $100\%$ & & $26\%$ & $24\%$ & \\
      with resnet instead of GRU                  & $20\%$ & $91\%$ & & $16\%$ & $18\%$ & \\
      \hline
    \end{tabular}
    \caption{Generalization of training error of Neural GPU is highly
      dependent on initialization seed. Table measures fraction seeds
      achieved error $<1\%$. Each entry of this matrix involves
      running experiments with $15$ different seeds on $13$ different
      tasks.}
    \label{fig:hyper}
\end{figure}

\begin{figure}[!h]
  \centering
  \includegraphics[width=.49\textwidth]{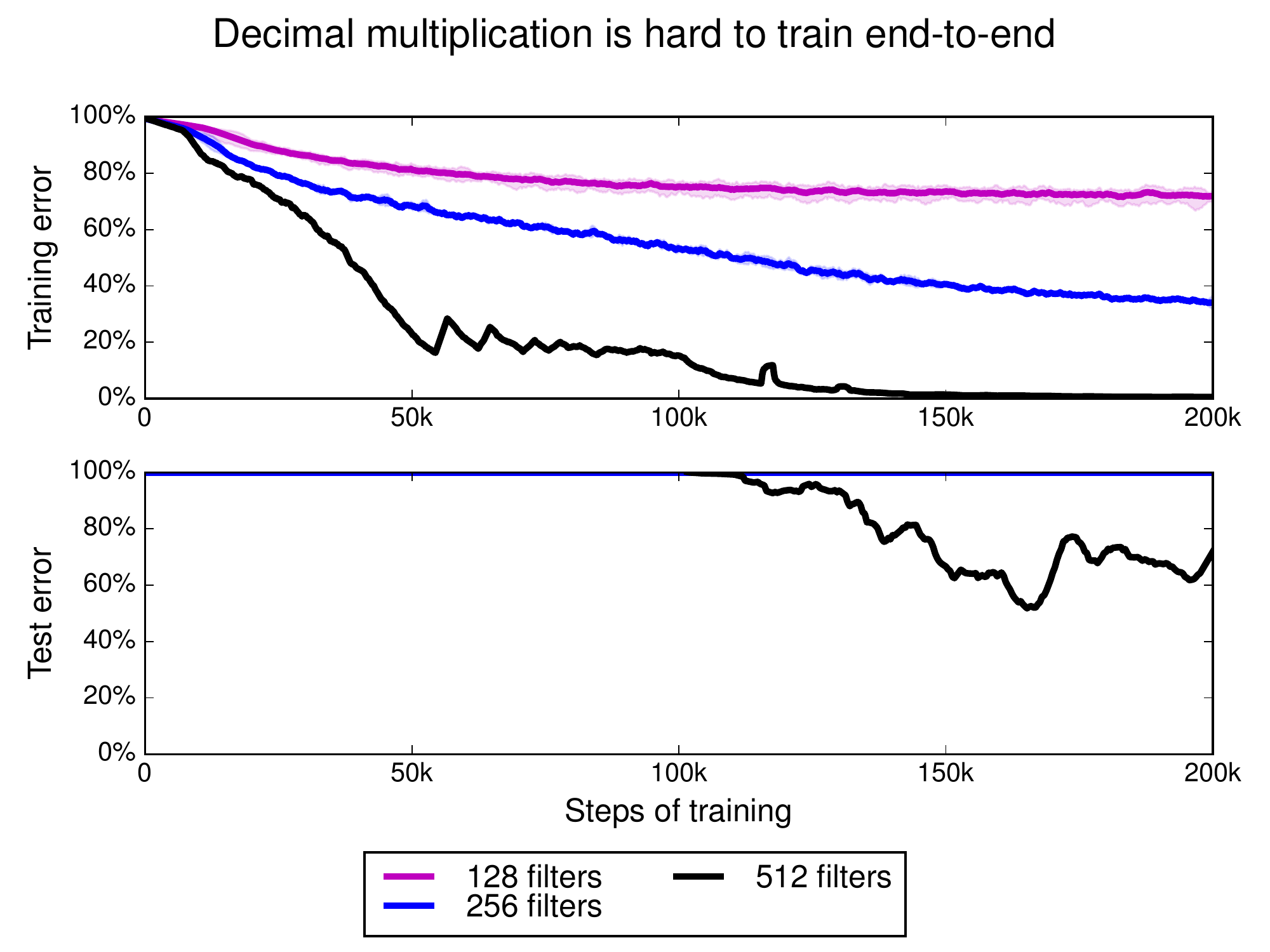}
  \includegraphics[width=.49\textwidth]{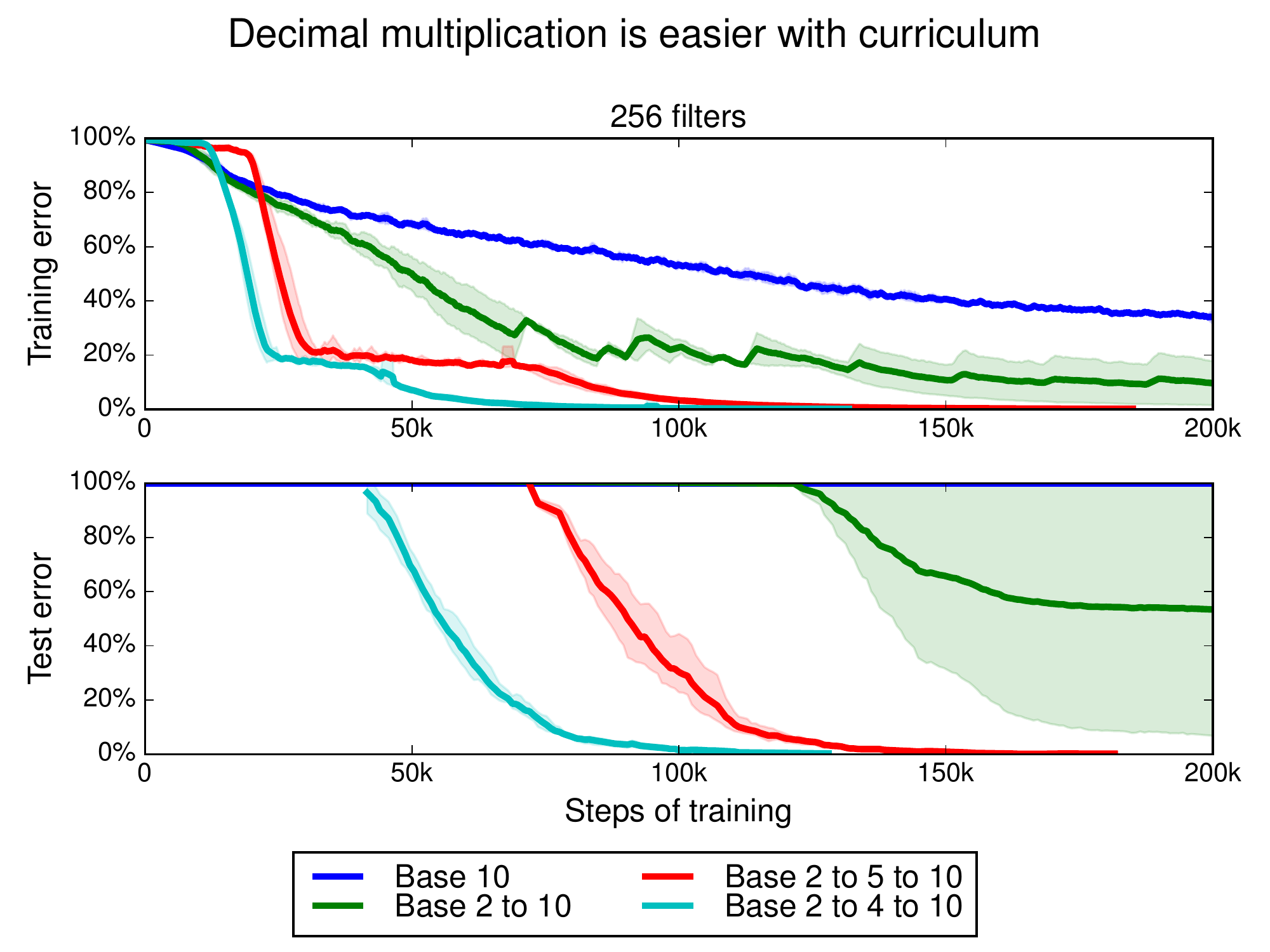}
  \includegraphics[width=.49\textwidth]{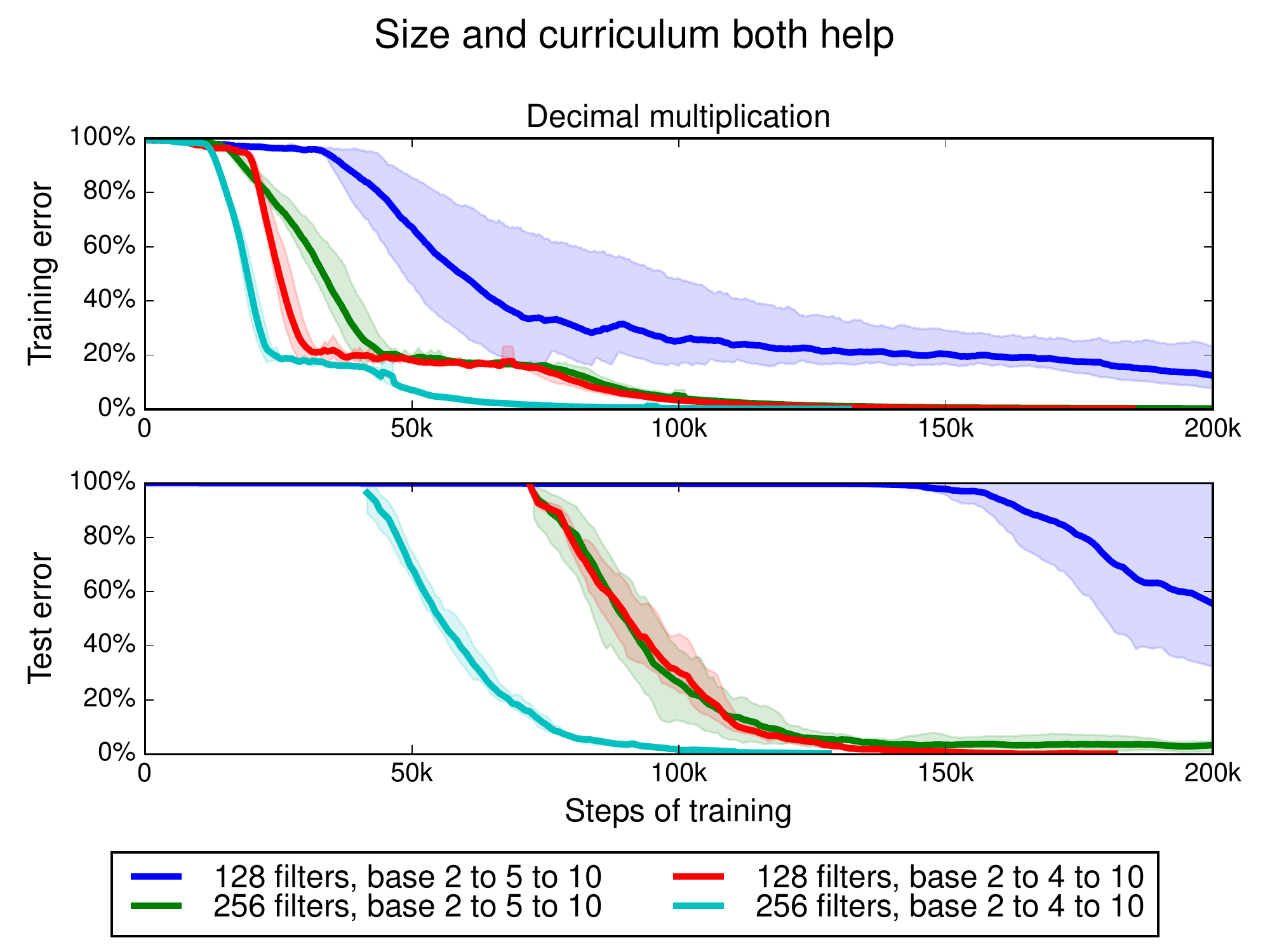}
  \caption{Training (top) and test error (bottom) on the decimal
    multiplication task.  (First) models of different sizes trained
    without curriculum. Only very large models can successfully
    optimize the training error, and none are able to generalize.
    (Second) variable-sized models trained with curriculum over
    different bases.  For instance, we find that a medium-sized model
    generalizes to much longer test cases on decimal multiplication if
    it is first trained on binary multiplication and then on
    quaternary multiplication, before being trained on decimal
    multiplication. (Third) the best results are achieved when we
    apply curriculum on models with large numbers of filters.
    We find that training a model first on binary numbers, then 4-ary,
    then decimal works significantly better than training it on binary
    numbers, then 5-ary, then decimal.  The difference is comparable
    to the difference between 128 and 256 filters.}
  \label{fig:decimal}
\end{figure}

Our experiments in \fig{hyper} show that the simplest way to increase
performance is to increase the size of the model.  The rows of
\tab{summary} consistently show that larger models are more likely to
generalize.  In contrast, smaller models are often unable to minimize
training error. For example, we were not able to achieve $0\%$
training error when models with $128$ and $256$ filters are trained on
decimal multiplication task (left \fig{decimal}), while the larger
models with $512$ filters are achieve $0\%$ training error (although
they do not generalize to long test inputs since curriculum was not
used).  It is not self-evident that larger model would generalize
better, but the empirical results are clear. We suspect that the
over-parameterization of the model makes the optimization task easier.

Running models with $512$ filters is challenging, because they do not
fit into the memory of the GPU on the problems that we have been
exploring.  In the implementation used for the experiments in this paper, a Neural GPU with $128$ filters does
not fit into GPU memory (12 GB) due to its large number of
activations.  To train larger models, we used several techniques for
reducing the model's memory footprint.  First, instead of unrolling
the graph, we used TensorFlow's symbolic \texttt{tf.while\_loop},
which can store intermediate activations on CPU thanks to the \texttt{swap memory}
option (implemented in tensorflow~\citep{abadi2015tensorflow}).
Second, we have decreased the batch size, which further lowered the memory
footprint.  A third way would be to use the methods described
by~\cite{martens2012training, gruslys2016memory}, however we have not
experimented with them.

\begin{table}[!h]
\tiny
\centering
\renewcommand{\arraystretch}{1.1}
\begin{tabular}{|l||c|c|c|}
  \hline
  \diagbox{Curriculum}{\# filters}  & 128  & 256  & 512 \\
  \hline
  \hline
  10  & \fails & \fails & \struggles \\
  {$2 \to 10$}  &  \fails & \struggles & \struggles \\
  {$2 \to 5 \to 10$}  &   \struggles & \works & \works \\
  {$2 \to 4 \to 10$  }& \works & \works & \works \\
  \hline
\end{tabular}
\caption{The table summarizes the setting where the Neural GPU
  succeeds to learn decimal multiplication.  It is essential to use a
  large model and to use a high-quality curriculum. The curriculum
  column indicates the base in which training took place; i.e. $2 \to
  4 \to 10$ means that we have started training in the binary base,
  then we moved to quaternary base, and finally to the decimal.}
\label{tab:summary}
\end{table}

In its current form, the Neural GPU cannot reliably learn to solve
hard tasks such as decimal multiplication from examples (although
sporadically it does).  We find that an elaborate curriculum is
necessary in order to learn such problems reliably, and to solve
harder tasks.  Our curriculum involves starting with short examples
first, and solving intermediate task on the way to solve the target
task.  For instance, a curriculum for decimal multiplication could
consist of first learning multiplication in base 2, then base 4, and
finally base 10.  The second plot in \fig{decimal} shows the training
and test error for models trained with curriculum.  The third plot in
\fig{decimal} presents results for different curriculum types and
model sizes.  \tab{summary} summarizes the success rates for different
model sizes, and curriculum.

\begin{figure}[!h]
  \centering

  \begin{minipage}{.48\linewidth}
  \includegraphics[width=\textwidth]{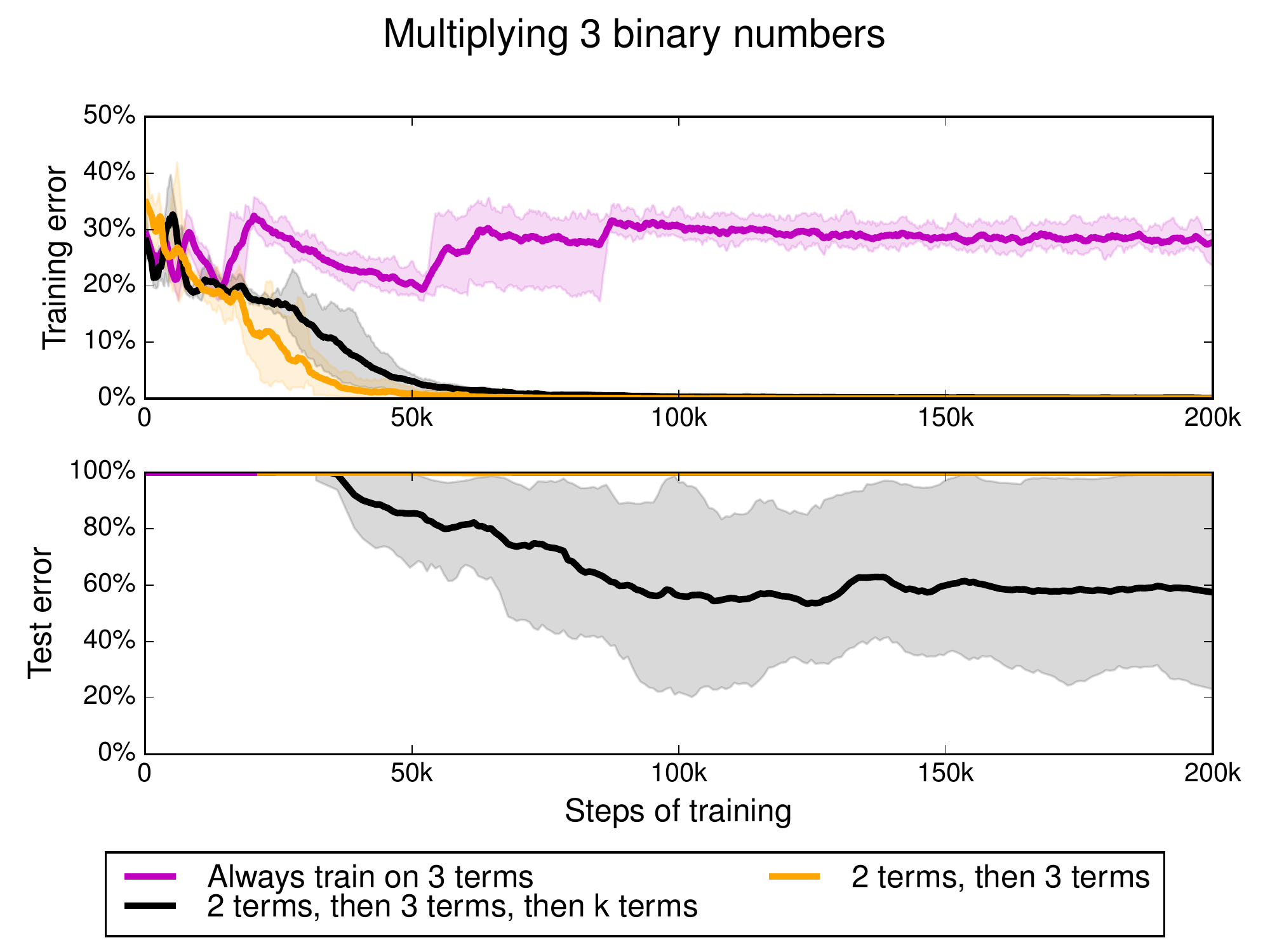}
  \end{minipage}
  \begin{minipage}{.48\linewidth}
    We look at the binary multiplication of 3 terms.  We find that we
    cannot learn this task directly even on training data.  However,
    if we start out learning binary multiplication of 2 terms, and
    then we switch to 3 terms, we get low training error, but the test
    error is 100\%.  However, if instead we train on the task of mulitplying
    $l$ terms, where
    $l$ is randomly chosen between $1$ and half the expression length,
    then we achieve low training error and non-trivial test error on the task
    of multiplying three terms.
  \end{minipage}
  \hspace{1cm}
  \caption{Influence of curriculum on 3-numbers multiplication task.}
    \label{fig:3bmul}
\end{figure}

We have obtained other notable results on the 3-numbers multiplication
task~(\fig{3bmul}).
We were not able to reach perfect error when
training on such this task without a curriculum. We found that training
on multiplication of 2 numbers and then moving to multiplication of 3
numbers improved performance.  However, such model doesn't
generalize. The best performance on the 3-numbers multiplication is
achieved when each training instance consists $l$ numbers, where
$l$ is randomly chosen between $1$ and $\lfloor \frac{n-1}{2}\rfloor$,
where $n$ is the length of a single number.

\begin{figure}
  \centering
  \begin{minipage}{.5\linewidth}
    \includegraphics[width=\textwidth]{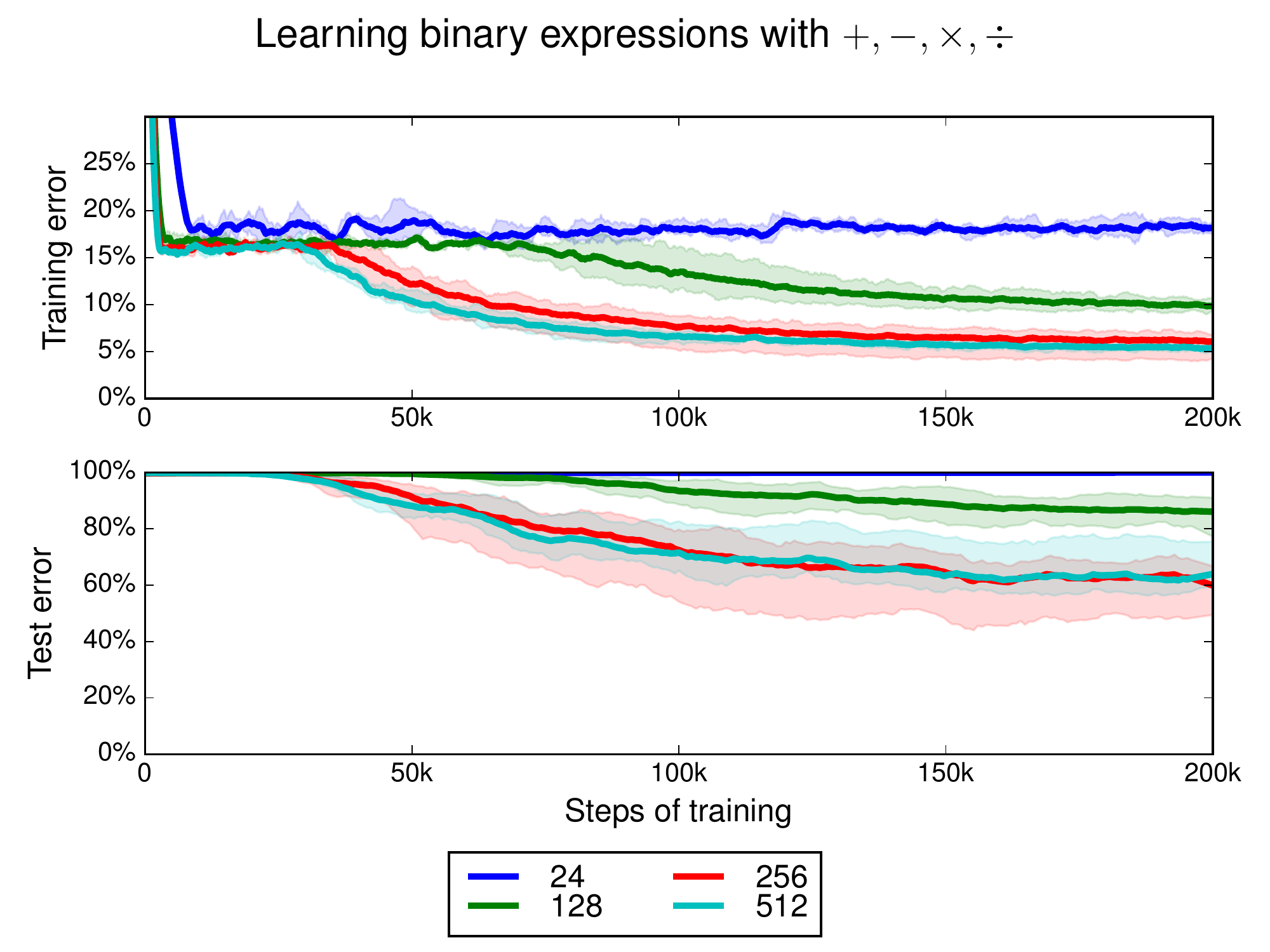}
  \end{minipage}
  \begin{minipage}{.45\textwidth}
    Each character has a 70\% chance of being a random digit, and a
    30\% chance of being a random operator, subject to operators not
    being adjacent and the result being nonnegative.  Division is
    interpreted as integer division.\\

Example input (on one line):
{\tiny
\begin{verbatim}
1101-0*11+100+0+111+1-000011-1*1110/1101001*1001
+0-10*11*00100/1111-011*1+010+1*00100010101001-0
00*1000110100/1/011000001+1*0/111-10/1/10*0*001*
1/001/11+0/010101+0+0*1+0011+01-0/00110+01*100+0
00/11-101
\end{verbatim}
}

Output: $\underbrace{\texttt{0}\dotsb\texttt{0}}_{189}\texttt{100011000000}$.
  \end{minipage}
  \caption{Task of learning binary arithmetic on multiple numbers simultaneously using the operators $+,-,\times,\div$.}
  \label{fig:bexpra}
\end{figure}

Another experiment is to train a model on sequences of arithmetic
operations with multiple numbers. We train on expressions of length
$41$, and test on expressions of length $201$.  When using the binary representation, our model
is able to correctly solve $30\%$ of length-201 expressions, where as always we
count success only when the whole number is predicted correctly.  For an
example of successful generalization, the model successfully evaluated
the expression
{\tiny\texttt{001110111001/1+10-10*0/1/1*111/10*010+0/010011-10/10101*0+
    010/1*00-110*1*0/1/101000- 00000+ 000-1+
    1-1011111*010-010/0011111011-010-1100-0/0010000*01*0010000+
    0111110+ 00001+10/10*111111111-10*10-1*11111+01}} (whose value is
{\tiny\texttt{10100001011}}).    \fig{bexpra} summarizes this result.

\section{Generalization}

Integer addition is a well defined algorithm, so knowledge of its operation is
sufficient to add arbitrary numbers. Our trained model generalize
perfectly to $>99\%$ of uniformly random test cases of much greater length (having a single
digit error in the whole $100$-digit number is considered an error).
However, we found that it fails on a much larger fraction of
``structured'' examples. For example, it fails on cases that involve
carrying a digit for many steps.

\begin{figure}
  \centering
  \includegraphics[width=0.49\textwidth]{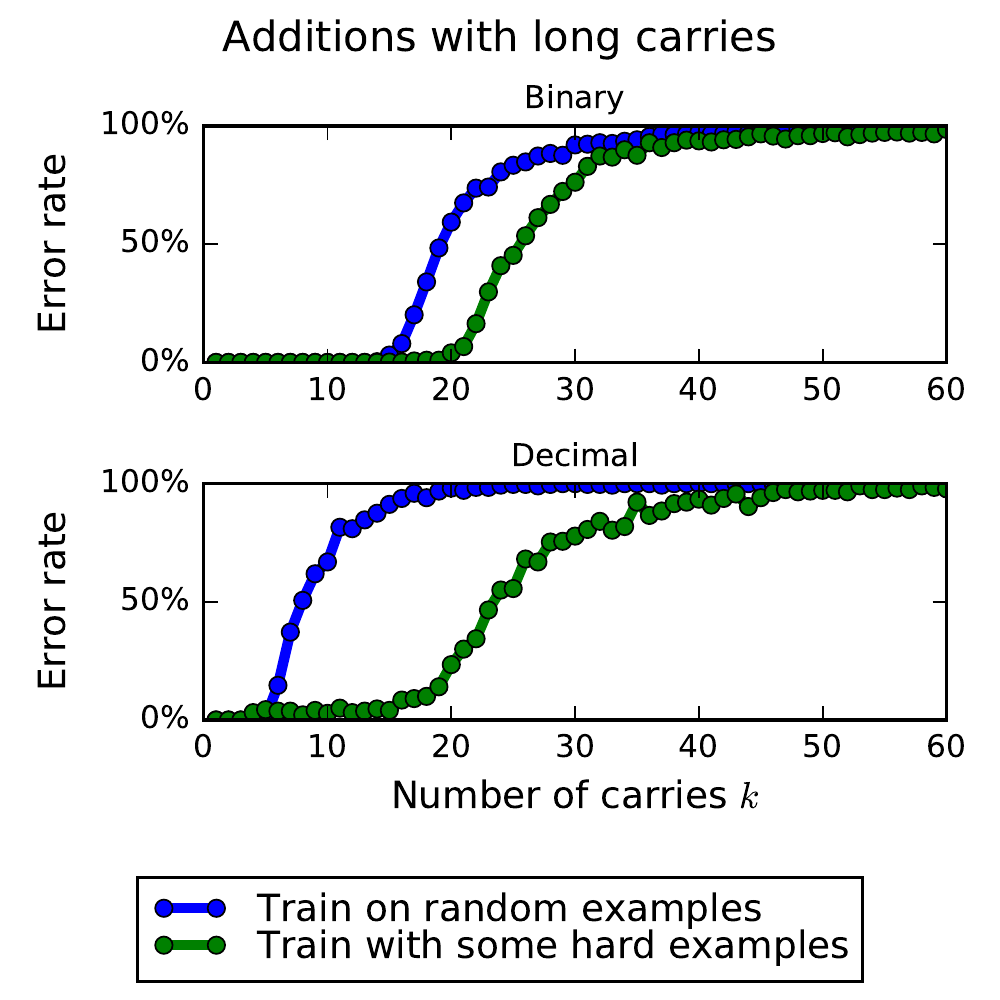}
  \includegraphics[width=0.49\textwidth]{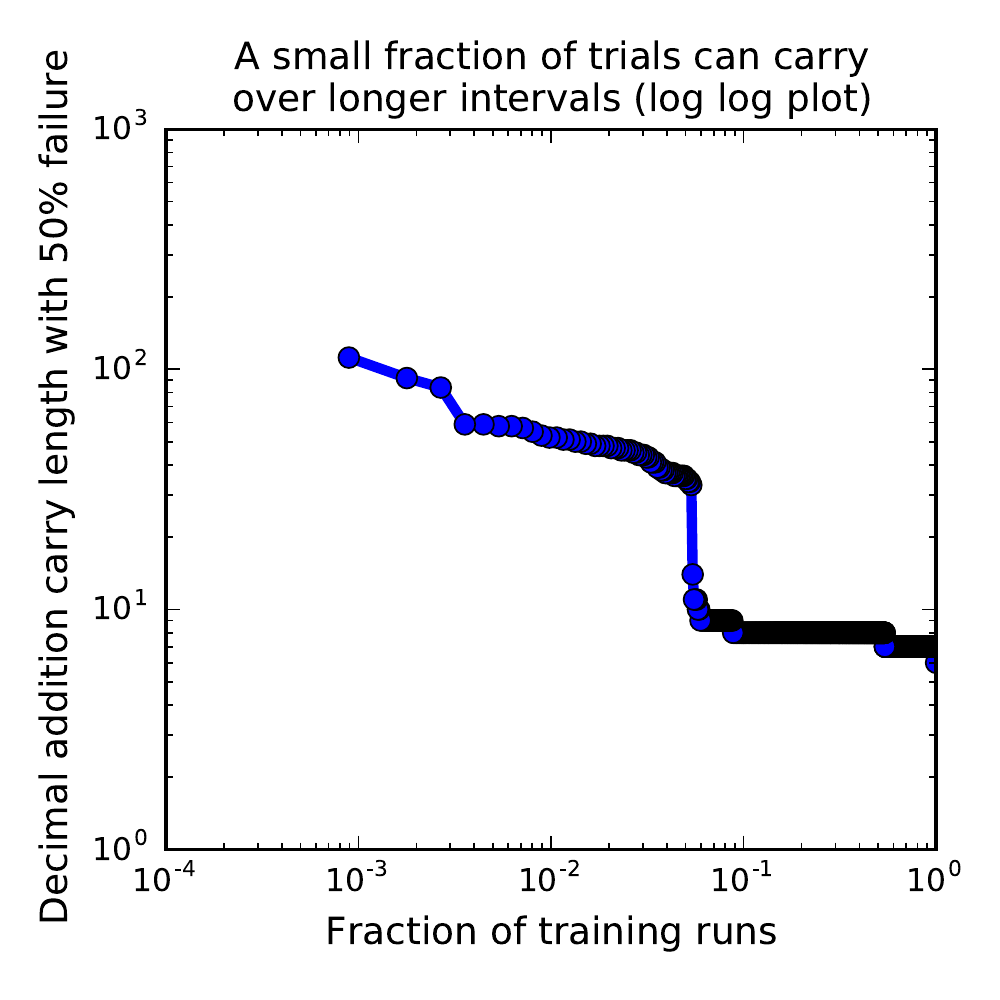}
  \caption{Left: while trained models for addition work on random
    inputs, they do not reliably work on inputs containing long carry
    sequences.  Right: occasional training runs yield models that
    carry over somewhat longer sequences.  }
  \label{fig:carries}
\end{figure}

The probability that a bit is to be carried $k$ times, on uniformly random
binary inputs, is $2^{-k}$.  Training and testing on random examples
will thus not notice failure to carry more than about $20$ bits. We
find that the model typically fails to generalize on test cases that
require more than $20$ steps of carry, as shown in \fig{carries}.
Similarly, with decimal addition, the chance of carrying $k$ digits is
$10^k$ and most trained models stop working reliably after 6 digits.

Introducing ``hard'' training instances with long carries does not
solve this issue.  Our training instances involve 20-digit numbers, so
they can only carry for 20 digits.  Indeed, we find that the trained
model will usually answer length-20 carry inputs correctly, but not
length-30.

The above behavior holds for most trained models, but as
in~\cite{KS15} we find that training the Neural GPU with different
random seeds will occasionally find models that generalize better.  We
trained a decimal addition model with 1121 different seeds, and
measure for each model the number of carries at which the error rate
crosses 50\%.  As shown on the right of \fig{carries}, the
distribution is bimodal: 94\% of models have a threshold between 7 and
9 digits, while almost the entire remainder (5.5\%) have a threshold
of at least length 33.  These ``successful'' models seem to follow
roughly a power law distribution: 5\% are at least 33, 1\% are at
least 52, and our best performing model worked up to a carry of length 112.  If
this behavior is indeed a power law, it is not a very promising way to
find highly-generalizing models: a rough extrapolation suggests
generalizing to 1000 digits would require 200 million trials.

We observe similar generalization issues with multiplication. A model
that has $<1\%$ error on two random $100$-digit long decimal examples
fails much more often on very symmetric numbers and on numbers that
requires to carry for many steps (we test it by verifying performance
on numbers that multiply to $b^{100}$ or $b^{100} - 1$).  It also failed
much more often on multiplying single digit numbers when they are prepended with
many zeros (\tab{prepended}); one random trained model failed on 38 of the
100 possible pairs of single-digit numbers.

\begin{table}[!h]
\tiny
\centering
\renewcommand{\arraystretch}{1.1}
  \begin{tabular}{|c|c|c|c|}
    \hline
    {\bf Left Operand} & {\bf Right Operand} & {\bf Observed value } & { \bf Is prediction correct ? } \\
    \hline
        \multicolumn{4}{|c|}{Single digit examples} \\
    \hline
    $1$ & $1$ & $1$ & \yes \\
    $2$ & $1$ & $2$ & \yes \\
    $2$ & $2$ & $4$ & \yes \\
    $2$ & $3$ & $6$ & \yes \\
    $2$ & $4$ & $8$ & \yes \\
    $3$ & $3$ & $6$ & \yes \\
    $3$ & $5$ & $15$ & \yes \\
    \hline
        \multicolumn{4}{|c|}{Single digit examples with prepended zeros} \\
    \hline
    $\underbrace{0 \cdots 0}_{99}1$ & $\underbrace{0 \cdots 0}_{99}1$ & $\underbrace{0\cdots 0}_{199}1$ & \yes \\
    $\underbrace{0 \cdots 0}_{99}2$ & $\underbrace{0 \cdots 0}_{99}1$ & $\underbrace{0\cdots 0}_{144}017666666666885\underbrace{0\cdots 0}_{40}07$ & \no \\
    $\underbrace{0 \cdots 0}_{99}2$ & $\underbrace{0 \cdots 0}_{99}2$ & $\underbrace{0\cdots 0}_{144}022333333333770\underbrace{0\cdots 0}_{40}14$ & \no \\
    $\underbrace{0 \cdots 0}_{99}2$ & $\underbrace{0 \cdots 0}_{99}3$ & $\underbrace{0\cdots 0}_{144}066000000000655\underbrace{0\cdots 0}_{40}21$ & \no \\
    $\underbrace{0 \cdots 0}_{99}2$ & $\underbrace{0 \cdots 0}_{99}4$ & $\underbrace{0\cdots 0}_{144}075166666667540\underbrace{0\cdots 0}_{40}28$ & \no \\
    $\underbrace{0 \cdots 0}_{99}3$ & $\underbrace{0 \cdots 0}_{99}3$ & $\underbrace{0\cdots 0}_{199}9$ & \yes \\
    $\underbrace{0 \cdots 0}_{99}3$ & $\underbrace{0 \cdots 0}_{99}5$ & $\underbrace{0\cdots 0}_{198}15$ & \yes \\
    & & & \\
    \hline
  \end{tabular}
  \caption{Neural GPU generalizes well to random $100$-digit examples. 
  However, it might fail to multiply single digit examples with large number of prepended zeros. Models trained on different seeds suffer on different examples with 
  prepended zeros. Therefore, problem is removable by model averaging. However, they all consistently fail on examples that require carrying for many steps.}
  \label{tab:prepended}
\end{table}

The multiplication model also fails on highly regular cases.  The outputs of the multiplication model are similar
to the correct answer, however it seems that the model made a mistake
in the middle of the computation.  For instance, we took two numbers
that multiply to $10^{180} - 1$. Such a number is full of
nines. However, model predicted: $10^{180} + 10^{103} - 1$
instead. The predicted number differs on $78$ positions, because it
has many zeros instead of nines; at some point, it starts to guess
that it is getting a carry.  Moreover, the model has trouble with very
structured expressions, such as $1111111111 \times 1111111111$.  It
fails for any number of ones above $8$.

\section{Global operation}

The Neural GPU is a cellular automaton, which is a Turing complete
computational model \citep{chopard1998cellular, wolfram1984cellular}.
However, the automaton is often computationally inefficient compared
to the von Neumann architecture.  It is difficult for a cellular automaton
to move data globally as the entirety of its computation operates
locally at each step.  We wanted to understand the importance of
globally moving data around for solving algorithmic tasks.

In principle, the Neural GPU could be made more powerful by adding a global
operation in each of its layers.
We have briefly tried to include an attention mechanism that has the ability
to easily shift rectangles of data to locations specified by the model,
but we were not able to improve empirical results. We skip these
experiments here, and instead investigate a simpler form of this question.

Given that the Neural GPU cannot easily move information across long
distances, it is conceivable that having both arguments concatenated
would hurt its performance.  We therefore experimented with several
input representations, as described below:
\begin{itemize}
    \item Padded: \texttt{12345+00067} 
    \item Unpadded: \texttt{12345+67} 
    \item Aligned:
      \begin{tabular}{r}
        \texttt{12345}\\
        \texttt{+00067}
      \end{tabular} 
\end{itemize}

\begin{figure}
  \centering
  \includegraphics[width=.6\textwidth]{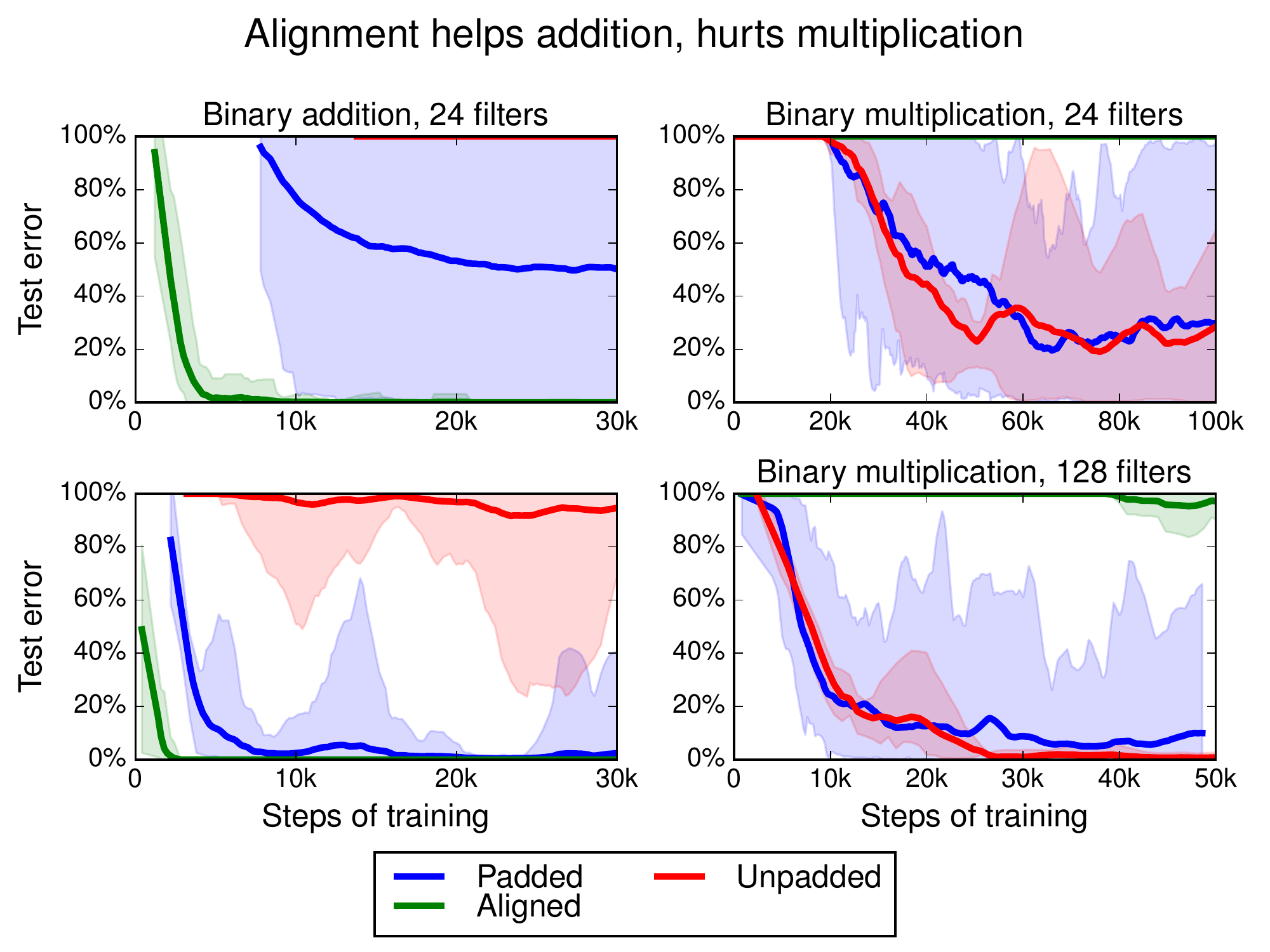}
  \caption{This figure shows the importance of input argument alignment on performance
    on the addition and the multiplication tasks. We find that alignment is
    beneficial to addition, but detrimental to multiplication.  Note that these
    experiments were conducted with a small neural GPU that had only 24 filters, so
    its generalization error is not as strong as that of the larger models.  However,
    the overall patterns tend to carry over from small models to large ones.}
  \label{fig:aligned}
\end{figure}

Could we manually align data in a way that helps to solve a task? This
could indicate that an architectural modification that performs a
global operation is needed.  And indeed, we found that the addition
task on aligned numbers has higher success rate than on concatenated
numbers, and addition on unpadded numbers is the hardest one
(\fig{aligned}, left).

However, we observe the opposite outcome for the multiplication task
(\fig{aligned}, right).  Empirically, aligning numbers for multiplication makes the
task more difficult.  These results suggest that an architectural
modification that makes it easy to move the data globally need not
provide an improvement on all problems.

\section{Conclusion}

In this paper, we investigated the generalization ability of the Neural GPU.
We have discovered that larger Neural GPUs generalize better, and provided
examples of several curriculums that made it possible for the Neural GPU to solve
tasks that it was not able to solve before.  Finally, we showed that its generalization
is incomplete, as while it has successfully generalized to longer inputs,
there still exist highly structured test cases that cause the model to fail.

It is desirable to develop learning methods that can learn algorithms that
achieve perfect generalization.  One way of moving forward is to investigate
ways in which the model can benefit from additional sources of information
that are not present in the task itself.

\section{Acknowledgment}
We wish to thank Rafal Jozefowicz for useful discussions and comments.

\bibliographystyle{iclr2016_conference}
\bibliography{refs}

\end{document}